\documentclass[letterpaper]{article} 
\usepackage{aaai23}  
\usepackage{times}  
\usepackage{helvet}  
\usepackage{courier}  
\usepackage[hyphens]{url}  
\usepackage{graphicx} 
\usepackage{natbib}  
\usepackage{caption} 
\usepackage{algorithm}
\usepackage{algorithmic}

\usepackage{times}
\usepackage{soul}
\usepackage{url}
\usepackage{multirow}
\usepackage{amsthm}
\usepackage{subcaption, booktabs}

\usepackage{newfloat}
\usepackage{listings}
\DeclareCaptionStyle{ruled}{labelfont=normalfont,labelsep=colon,strut=off} 
\floatstyle{ruled}
\newfloat{listing}{tb}{lst}{}
\floatname{listing}{Listing}
\title{A Study of Slang Representation Methods}
\author{
Aravinda Kolla$^{1}$,
Filip Ilievski$^{1}$,
Hông-Ân Sandlin$^{2}$, and
Alain Mermoud$^{2}$
}
\affiliations{
$^1$University of Southern California, Information Sciences Institute, USA\\
$^2$Cyber-Defence Campus, armasuisse Science and Technology, Switzerland\\
\{kollaa, ilievski\}@isi.edu,
\{hongan.sandlin, alain.mermoud\}@ar.admin.ch
}

\begin{document}

\maketitle

\begin{abstract}

Considering the large amount of content created online by the minute, slang-aware automatic tools are critically needed to promote social good, and assist policymakers and moderators in restricting the spread of offensive language, abuse, and hate speech. Despite the success of large language models and the spontaneous emergence of slang dictionaries, it is unclear how far their combination goes in terms of slang understanding for downstream social good tasks. In this paper, we provide a framework to study different combinations of representation learning models and knowledge resources for a variety of downstream tasks that rely on slang understanding. Our experiments show the superiority of models that have been pre-trained on social media data, while the impact of dictionaries is positive only for static word embeddings. Our error analysis identifies core challenges for slang representation learning, including out-of-vocabulary words, polysemy, variance, and annotation disagreements, which can be traced to characteristics of slang as a quickly evolving and highly subjective language.
\end{abstract}

\section{Introduction}


The UN Sustainable Development Goals~\cite{biermann2017global} emphasize the importance of gender equality, peace, and justice. Initiatives that envision the role of AI for social good~\cite{tomavsev2020ai} provide guidelines for supporting these goals through practical measures. A key application of AI is building tools that assist social media policymakers and moderators to restrict the spread of offensive language, abuse, and hate speech, e.g., sexism and misogyny, which are still prevalent all over the globe~\cite{khan_2021}. 
A recent study~\cite{delisle2019large} reveals worrying patterns of online abuse, estimating 1.1 million toxic tweets sent to women over one year. Another study by \citeauthor{mamie2021anti} demonstrates how ``the Manosphere'', a conglomerate of men-centered online communities, may serve as a gateway to far-right movements. Web content moderation policies, or the lack thereof, can have serious implications on individuals, groups, and society as a whole.
On the one hand, content moderators may react late, inconsistently, or unfairly, thus angering users~\cite{Habib_Nithyanand_2022}, as well as contributing to reinforcing and exacerbating conspiratorial narratives \cite{chen2021neutral}.
On the other hand, minimal content moderation may permit coordinated influence operations~\cite{diresta2019potemkin} or enable the spontaneous formation of toxic and dangerous communities~\cite{mamie2021anti,khan_2021}.

While (computational) linguistics has typically focused on formal documents, like books, and curated text corpora, like Wikipedia, language on the internet is informal, dynamic, ever-evolving, and bottom-up~\cite{mcculloch2020because}. Thus, automatic content analysis tools designed to assist social media policymakers and moderators need to possess the ability to understand informal language, i.e., \textit{slang}. Slang, defined as ``a peculiar kind of vagabond language, always hanging on the outskirts of legitimate speech, but continually straying or forcing its way into the most respectable company''~\cite{greenough1901words}, has long been of interest to linguists and social historians~\cite{adams2012slang,partridge2015slang,detectslang}. With the velocity and the volume of informal content on the Web, the notion of slang has recently broadened~\cite{mcculloch2020because}: slang has moved beyond its traditional categories, like school, intergenerational, and intragenerational slang~\cite{adams2012slang}, to loosely-defined communities and subcultures, like QAnon followers or men-centered online communities~\cite{mamie2021anti}.

How can we build representation learning methods that are capable of understanding slang? Can we apply or adapt large language models (LMs) for this purpose? Do dictionaries of slang or large social media datasets provide an effective opening for building specialized slang methods? Recognizing that most of the current LMs have been trained on formal English datasets that fail to capture the nuances of the social media language, prior work has devised tasks that test their abilities~\cite{2020-semeval,rosenthal-etal-2017-semeval}, methods that fine-tune them on benchmark-specific data~\cite{bertweet}, and resources that define and describe a comprehensive collection of slang words~\cite{emo-love}. It is unclear to what extent recent models and knowledge sources can understand slang, and what are the key limitations that affect their reasoning on slang-centric downstream tasks.




\begin{figure*}[!t]
    \centering
    \includegraphics[width=0.8\linewidth]{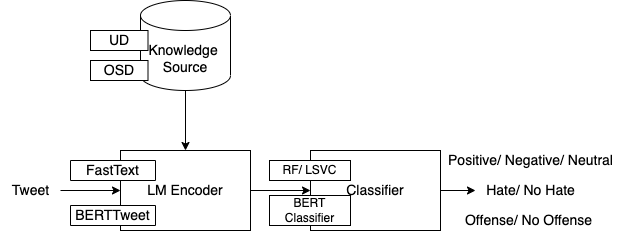}
    \caption{Overview of the Framework. RF is Random Forest Classifier and LSVC is Linear SVC classifier. UD is Urban Dictionary and OSD is online slang dictionary.}
    \label{fig:overview}
\end{figure*}

In this paper, we study the ability of combinations of language models and knowledge sources to comprehend slang in different tasks involving offense, hate, and abuse on social media. We analyze their behavior quantitatively and qualitatively, in order to surface the chief weaknesses of these models in terms of slang understanding and point to potential solutions going forward.
We make the following contributions:
\begin{enumerate}
    \item We design and implement a framework for slang representation learning methods. The framework integrates language model encoders, knowledge sources, and evaluation tasks.  
    \item We design experiments to answer key research questions empirically. We evaluate if existing FastText models trained on slang data answer key research questions. Our experiments with three models, two knowledge sources, and three tasks reveal differences between the ability of models to capture slang, as well as inherent challenges that require innovative approaches. 
    \item We reflect on our findings, connecting them to the characteristics of slang as a quickly evolving language, and as a subjective and circumstantial phenomenon causing ambiguity in downstream tasks.
\end{enumerate}

We make our code and data available at \url{https://github.com/usc-isi-i2/slang-representation-learning}.

\section{Slang Representation Framework}
\label{sec:framework}



Our study is based on a framework that combines different language models, knowledge sources, and evaluation tasks (Figure \ref{fig:overview}). We describe the framework's components in turn.

\subsection{Models} 

\textbf{FastText} FastText embeddings represent each word as a bag of character n-grams, which helps in overcoming the limitation of models that ignore the morphology of words \cite{fasttext}.
In \cite{wilson-etal-2020-urban}, the authors train a FastText model on the Urban Dictionary dataset,\footnote{https://www.urbandictionary.com/} and evaluate these embeddings on both intrinsic tasks: Semantic Similarity and Clustering, and extrinsic tasks: Sentiment Analysis and Sarcasm Detection. In this work, we train FastText embeddings using the skip-gram architecture on Urban Dictionary and Online Slang Dictionary data.  
To make predictions with the FastText embeddings on discriminative tasks, we use Random Forest and Linear SVC classifiers on all the evaluation datasets. 



\begin{table*}[!t]
    \centering
    \caption{Examples from the evaluation datasets with corresponding labels. }
    \label{tab:examples}
    \begin{tabular}{|p{3.2cm}|p{11cm}|p{1.5cm}|}
    \hline
         Evaluation Task & Examples & Label  \\
         \hline
        \multirow{3}{*}{Sentiment Analysis}& so much love for this woman, ughh \#arianagrande & positive  \\
        & Give it a listen! Raw project from one of the realest \#blacklivesmatter & neutral \\
        & Cannot believe people are attempting to sell a \$50 toy for 6x the price! Unbelievable! \#Hatchimals \#peoplesuck & negative \\
        \hline
        \multirow{2}{*}{Hate Speech Detection} &  There are NO INNOCENT people in detention centres \#SendThemBack 
        & 1\\
        &Libyan coast guard rescues some 160 \#Europe-bound migrants &0 \\
        \hline
        \multirow{2}{*}{Offense Detection} & You forget Eastern Europeans, who are the worst 
        &1\\
        & Dude.. Why r u trolling Ash rock z?  He is our guy.. Just in a hurry to see bjp govt in state.& 0\\
        \hline
    \end{tabular}
\end{table*}
    
\textbf{BERT}
     Bidirectional Encoder Representations from Transformers (BERT) \cite{bert} is a large language model that outputs textual embeddings conditioned on both left and right context. BERT models can be pretrained in an unsupervised manner to get general language representations that can be used for downstream tasks. BERT \cite{bert} is trained in the formal English language from Wikipedia. As the BERT model has been largely trained on formal language in Wikipedia and Book Corpus, it can be anticipated to handle social media data suboptimally. Thus, we also experiment with 
     training a BERT model from scratch on slang-specific datasets.
     
\textbf{BERTTweet}
    As we expect that a BERT model trained on social media data is better equipped to capture slang language, we also experiment with the BERTTweet \cite{bertweet} model. 
    BERTTweet is pretrained on a large corpus of Twitter data containing around 873M tweets ranging from 2010 to 2019. BERTTweet can be expected to capture the commonly occurring slang words, for example, the word "lol" would be correctly tokenized as "lol". 


\begin{table}[!t]
    \centering
    \caption{Data Partitions of the Evaluation Datasets}
    \label{tab:data_part}
    \begin{tabular}{|c|c|c|}
    \hline
         Dataset & Train Size & Test Size \\
         \hline
          Sentiment Analysis & 32084 &  8021 \\
          \hline
          HateEval & 8000 & 2000 \\
          \hline
          OffenseEval & 10592 &  2648 \\
          \hline
    \end{tabular}

\end{table}




\subsection{Knowledge Sources}
\textbf{Urban Dictionary (UD)} has been a standard source of slang in many works such as \cite{wilson-etal-2020-urban}, \cite{slangsd}, and \cite{slangnet}. Urban Dictionary is a crowd-sourced collection of slang words along with their definitions and usage examples, containing over two million entries. Urban Dictionary, however, has its own demerits. The dictionary's voting system and its low threshold to include the new content result in noisy and opinionated entries. A detailed exploratory study of the urban dictionary  is presented in \cite{emo-love}. For training our models, we only work with the usage examples from the Urban Dictionary. There are two usage examples for each word on average resulting in around 4 million data entries.

\textbf{Online Slang Dictionary (OSD)} contains fewer entries in comparison to the Urban Dictionary, around 15k in total.\footnote{\url{http://onlineslangdictionary.com/}} OSD contains words/phrases, their meaning, and their usage in sentences. We extract the 12k entries from OSD that  contain usage examples. 

 In addition to the Urban Dictionary and the Online Slang Dictionary, we also explore Green's Dictionary of Slang.\footnote{\url{https://greensdictofslang.com/}} Green's Dictionary of Slang describes the origin of words and phrases. In our context, this data source is not useful as it does not contain any usage examples and hence is not included in the reported experiments. We also explored purely slang-based Twitter corpora from \cite{slangvolution}. We experimented by retraining the BERTTweet model on this dataset to address the frequency of the slang words issue in the slang dictionaries dataset. However, the retrained model reported lower accuracy than that of the base BERTTweet model. 

\subsection{Classifiers}
We select commonly used classifiers such as Random Forest and Linear SVC to classify the FastText embeddings of the tweets in each of the evaluation tasks. Random Forest is an ensemble learning-based machine learning model that constructs multiple decision trees on the training data. The classification prediction of Random Forest is the class predicted by most decision trees. Linear SVC is a support vector machine-based classification model. This model tries to find the best hyperplane that maximizes the distance between the samples from different classes.

The first layers of the BERT model learn generic linguistic patterns and the last few layers learn task-specific patterns. For downstream tasks such as classification, the first layers are frozen and the last layers are trained.

\section{Experimental Setup}
\label{sec:exp}
\subsection{Model Training Details}
\textbf{Training setup} We train the FastText models in an unsupervised manner on UD and OSD data. To train the FastText models, we use the standard skip-gram architecture and fine-tune it over the training data of each task. 
We report the results from Random Forest as it performs slightly better than SVC, noting that the difference is negligible.
The BERT model is pre-trained from scratch on the UD and OSD data with a masked language modeling objective. The BERTTweet model is re-trained on UD and OSD data and is fine-tuned on HateEval and OffenseEval datasets. BERT and BERTTweet are finetuned for the Sequence Classification task on the evaluation datasets.

For UD, we reuse the dataset provided by \cite{wilson-etal-2020-urban} containing over 4 million usage entries of slang words and their usage examples.
For OSD, we manually extract and clean the data entries resulting in a dataset of size 12k.



\textbf{Parameters}
We train the FastText model for 10 epochs with an embedding dimension of 300. We use 100 estimators for random forest classifier and l2 penalty loss for Linear SVC.
We finetune the BERTTweet model using the Adam optimizer with a learning rate of 1e-6 and cross-entropy loss for 5 epochs. We retrain the model only for 2 epochs as increasing the number of retraining epochs caused the model to distort its original knowledge acquired from the Twitter dataset by randomising the initial weights, leading to worse performance on the downstream tasks.

\subsection{Evaluation} \label{section:eval}

\textbf{Evaluation tasks} We evaluate our models on three classification tasks: Sentiment Analysis, Hate Speech Detection, and Offense Detection. We select these tasks because slang terminology is central to comprehension of hateful, emotional, and offensive content in social media communication.

\textbf{Data} For \textit{sentiment analysis}, we use the SemEval 2017 task 4 \cite{rosenthal-etal-2017-semeval} dataset containing 50k training entries and 12k testing entries.
SemEval 2019 \cite{basile-etal-2019-semeval} task 5 focuses on the detection of \textit{hate speech} against women and immigrants with a dataset of 13k tweets in English. 
SemEval 2020 \cite{2020-semeval} task 12 describes the task of \textit{offense detection} in social media with a dataset containing  15k entries. We show examples for each task with their corresponding labels in Table~\ref{tab:examples}.
We partition the evaluation datasets by randomly holding out 20\% of each dataset as validation dataset, and using the rest for training. Statistics per dataset are provided in Table~\ref{tab:data_part}.

We evaluate our models by using the customary metrics of precision, recall, and F1-score.



\begin{table*}[!t]
\centering
    \caption{Model performance evaluated on Sentiment Analysis, Hate speech and Offense detection.}
    \label{tab:results}
    \begin{tabular}{|c|c|c|c|c|}
    \hline
    Dataset & Model & Source & Accuracy & F1 score\\
    \hline
\multirow{4}{*}{Sentiment Analysis}  & FastText & - & 0.620 &\bf 0.710 \\
&  FastText & UD & 0.644 & 0.636 \\ 
& FastText & UD+OSD  & 0.650 & 0.642 \\
& BERT & - & 0.632 & 0.558 \\
& BERT & UD & 0.696& 0.648 \\
& BERTTweet & - &   \bf 0.704 & 0.674 \\
& BERTTweet & UD+OSD & 0.701 & 0.650 \\
    \hline
    \multirow{4}{*}{HateEval}  & FastText & - & 0.660 & 0.620 \\
&  FastText & UD & 0.670 & 0.620 \\ 
& FastText & UD+OSD  & 0.710 & 0.680 \\ 
& BERT & - & 0.774 & 0.743 \\
& BERT & UD & 0.795 & 0.744 \\ 
& BERTTweet & - &   0.824 & 0.802 \\
& BERTTweet & UD+OSD & \bf 0.856 & \bf 0.814 \\
\hline
\multirow{4}{*}{OffenseEval}  & FastText & - & 0.710 & 0.550 \\
&  FastText & UD & 0.720 & 0.570 \\ 
& FastText & UD+OSD  & 0.710 & 0.540 \\ 
& BERT & - & \bf 0.778 & 0.495  \\
& BERT & UD & 0.755 & 0.461 \\
& BERTTweet & - & 0.797 & \bf 0.675 \\
& BERTTweet & UD+OSD & 0.797 & \bf 0.675 \\
\hline
\end{tabular}

\end{table*}

\subsection{Research Questions}

\begin{enumerate}
    \item \textbf{Do resulting models understand slang?} Large language models often perform well on the downstream classification tasks~\cite{devlin-etal-2019-bert,brown2020language}. But can these models comprehend the quickly evolving slang in social media platforms and beyond, and can they leverage the slang terminology to detect inappropriate better? To answer these questions, we apply our framework and observe the performance of various combinations of language models and knowledge sources per task.
    \item \textbf{Which language model is best equipped for understanding slang?} We compare static and contextual models, and we compare models that have been tuned to social media data to those that have not. We compare their performance and their qualitative behavior.
    \item \textbf{Which knowledge source provides more useful knowledge for adapting models?} We evaluate various knowledge sources to see which source best helps the models capture the domain. Here, we compare two slang dictionaries with different sizes and content types against the vanilla models without direct slang source adaptation.
    \item \textbf{Which cases are difficult for our models?} We closely examine the failure cases and hypothesize the potential causes for the model's erratic behavior. We connect these failure categories intuitively to architectural and training decisions in the models and our framework.
\end{enumerate}

\section{Results}
\label{sec:results}

\subsection{Main Results}

We show the results of the different combinations of models and knowledge sources in Table \ref{tab:results}.
The BERTTweet model performs best in all the evaluation tasks. This is intuitive, as the model trained on tweets can be expected to capture the social media domain best and can give the best results on the evaluation datasets as they are also based on Twitter. This hypothesis is verified to be true by the results of the evaluation datasets. To confirm that the performance is owed to better coverage of slang, we test BERTTweet model manually on random sentences with slang words masked. We observe that this model has a relatively good command of words that are used as both slang and non-slang terms, e.g.,
for the sentence: \textit{This place is amazing. It is [MASK]}, the [MASK] is predicted as "awesome". 

\begin{table}[!t]
    \centering
    \caption{Top 5 nearest neighbours for FastText trained on Wikipedia and on slang dictionaries.}
    \label{tab:ft_nn}
    \begin{tabular}{c c c}
         Word & Wikipedia & UD+OSD   \\
         \hline
         \multirow{5}{*}{lol} & kidding& roflolmao \\
         & hahaha& rofllmao\\
         & yeah& rofl \\
         &hahahaha& haha \\
         &thats& lolk \\
         \hline
         \multirow{5}{*}{nvm} & nvmc& bvm \\
         & nvmd& tvm\\
         & nvme& mvm \\
         &nvs& nevermind \\
         &nvmfs& nvr \\
         \hline
                  
    \end{tabular}

\end{table}

We also observe that the BERTTweet model trained on the initial Twitter dataset and retrained on the Urban Dictionary data does not show major improvements. This can be attributed to the low frequency of the slang words in the UD+OSD dataset, where each slang word occurs at most three-four times in the dataset. If the slang word is not commonly occurring in the pretrained tweet dataset, then the word is not captured by the BERT or BERTTweet models.

The FastText model trained only on usage examples from the UD or OSD data performs better than the baseline models for each task. These embeddings also capture the relationship between various slang words. Upon examining the top 10 nearest neighbors of commonly occurring slang words in the evaluation dataset, as shown in Table \ref{tab:ft_nn}, we obtain the words ``haha'', ``kidding'', and ``rofl'' as nearest neighbors for ``lol''.
They are also able to capture the abbreviations in some cases: "nevermind" is the nearest neighbor to its abbreviation "nvm", while "wazup" is the nearest neighbor to the phrase "what is up".  

The BERT uncased model trained on Wikipedia data gives better results on the evaluation datasets than the FastText models. But on a closer examination of the tokenization of the slang words in BERT, they are tokenized incorrectly. For example, "lol" is tokenized as "lo" and "l", "slut" is tokenized as "s" and "lut" and so on, indicating that the model tokenizer fails to capture the slang domain, i.e., BERT often treats slang words as out-of-vocabulary terms.
The BERT model trained on UD also fails to capture social media language. This can be due to the low frequency of slang words in the dataset. When this pretrained BERT model is tested on random sentences with the slang words masked, the model gives out nonsensical or no results. For the sentence: \textit{This place is amazing. It is [MASK]}, the prediction for [MASK] is ".". For the sentence, \textit{I got very angry at her. In the moment, i [MASK]-slapped her in my head.}, the model predicts the masked value as "and". These results show that the model is unable to learn the context of the slang words. In terms of overall accuracy, this model yields similar results as the baseline BERT model on the evaluation datasets. 

To overcome the incorrect tokenization of slang words by the BERT-based models, we also deliberately extend the BERTTweet tokenizer vocabulary with the slang words from the Online Slang Dictionary. This extension leads to an appropriate tokenization of the slang words, however, the model performance does not improve. The results of the extended model are shown in Table \ref{tab:exbert}.

\begin{table}[!t]
    \centering
    \caption{Evaluation of BERTTweet with an extended vocabulary.}
    \label{tab:exbert}
    \begin{tabular}{|p{2cm}|p{1cm}|p{1.2cm}|p{1.1cm}|p{1cm}|}
    \hline
    Dataset & Acc & Precision & Recall & F1 \\
    \hline
    
         Sentiment Analysis & 0.685 & 0.581 & 0.633 & 0.606 \\
         HateEval & 0.794 & 0.733 & 0.793 & 0.762  \\
         OffenseEval & 0.758 & 0.620 & 0.652 & 0.635\\
         \hline
    \end{tabular}
\end{table}

\subsection{Error Analysis}

\textit{Which cases are difficult for our models?} Through qualitative error analysis, we observe the misclassification occurs primarily due to five factors:
\begin{enumerate}
    \item \textbf{Incorrect tokenization of infrequent terms.} Due to the low frequency of the slang words in the dataset, the BERT models do not tokenize the slang words as expected. Novel tokens such as ``snatched'' and ``bae'' are typically split into subtokens, which may influence their classification in a negative way, as they contain key information that is missed by the model.
    \item \textbf{Tweets consisting mostly of URLs are misclassified.} This is because the content of the URL is not known without resolving it with an HTTP request. For example, the tweet \textit{``Austria proposes sending troops abroad to stop migrant movement https://t.co/cnbxbFYdBU, Immigration Jihad in action https://t.co/VuFN3DktJ7''} is difficult to classify without including the contents of the URL as additional context.
    \item \textbf{Polysemy between slang and non-slang words} such as "sick" and "cold" is difficult to cover for our models. For example, the word ``sick'' can be used with a negative, formal language, connotation: "I am feeling very sick", or a positive slang connotation "These beats are sick". Similarly, words like ``woke'' may be positive or negative depending on the context and the perspective of the author. Namely, a positive connotation of the term is used in the sense of one being aware and proactive about current affairs, whereas in a negative sense woke is an oxymoron about people that express opinions about any mainstream topic in an attempt to gain attention. The polysemy and the contextual dependency makes it difficult for the models to capture the intent the slang words are associated with.
    \item \textbf{Variance in spelling is non-trivial to handle for embedding models}. A common feature of a non-stabilized language, like slang or a dialect, is using a novel spelling of the word to emphasize the tone of the word or express emotion. An example is the infamous \textit{``heyyyy''} that turned into a meme. Additionally, given that the language is informal, there is often no single correct way to spell a word: "wazzzup", "wassup", "wassssssup", "wasup", and "sup" are all acceptable slang expressions.    
    
\end{enumerate}


\section{Discussion}
\label{sec:discussion}

In summary, we observe that models that are trained on large-scale social media data are most capable of understanding slang in downstream tasks. Slang-specific sources are 
mainly beneficial for static models that have been trained on Wikipedia corpora before. While out-of-vocabulary terms represent a key challenge for contextual models like BERT and BERTTweet, adapting these models or their tokenizers does not manifest in better performance in downstream reasoning tasks. Further issues relating to polysemy, variance in spelling, and annotation disagreements point to characteristics of the phenomenon of slang as a quickly evolving and subjective language, which we discuss next. 

\textbf{Slang as a quickly evolving language} Internet linguistics accelerates the inherent property of language to evolve over time. While the factors that influence language evolution are still hypothesized~\cite{kolodny2018evolution}, it is apparent that this evolution is largely accelerated with the emergence of Internet linguistics~\cite{mcculloch2020because}. The language on the Internet, expressed through novel forms including tweets and memes, leads to the quick evolution of expressions and the spreading of novel slang terms within communities, eventually forming an Internet folklore. On the one hand, this motivates the need for integrated AI solutions that mix many forms of Internet data: cultural tropes,\footnote{E.g., \url{https://tvtropes.org/}} memes,\footnote{E.g., \url{https://knowyourmeme.com/}} UD, Hatebase~\cite{boyd2022research}, and other forms of usage data. On the other hand, it is possible that slang can adapt very quickly to moderation and thus thwart this type of approach. In fact, it seems that slang and some memes are already specifically designed as euphemisms in order to "fly under the radar" of censorship, such as \textit{Pepe the Frog}.\footnote{\url{https://en.wikipedia.org/wiki/Pepe_the_Frog}}
From this perspective, it would be interesting to measure the influence/endogeneity of moderation and censorship on slang, the speed of adaptation, and the real efficiency of moderation to stop the diffusion of inappropriate content. 

\textbf{Slang as a subjective phenomenon} Tasks that involve hate speech or offense detection are inherently subjective and depend on the annotator's demographics, knowledge, and experience. Prior work reports that disagreement for sentiment analysis ranges between 40–60\% for low-quality annotations, and between 25–35\% even for high-quality annotations~\cite{kenyon2018sentiment}. Rather than ignoring the disagreement and evaluating on the majority label, a more sophisticated idea is to train and evaluate AI methods that can handle and predict human disagreement~\cite{kralj2022handling}. As suggested in~\cite{kocon2021offensive}, personalized models can be trained to detect and mimic individual or community profiles, thus treating the disagreement as a signal rather than noise. These models can be trained to explicitly model the psychological traits of the individual or group of annotators, inspired by the approach in \cite{bahgat2022liwc}.

\section{Related Work}
\cite{slangsd} created a sentiment dictionary for slang words for sentiment analysis of social media data. \cite{slangnet} built a WordNet-like resource for slang words and neologisms and the efficacy of this resource was evaluated over Word Sense Disambiguation algorithms for English social media data. 
\cite{wilson-etal-2020-urban} explored generating slang embeddings for Urban Dictionary data using the FastText framework. These embeddings were evaluated on a variety of tasks such as Sentiment Analysis and Sarcasm Detection. \cite{detectslang} attempts to detect and identify slang in Twitter data by using LSTM-based networks with feature boosting. 
Automated tools that capture slang have been attempted to associate Web content (on social media) to the psychological traits of the writer~\cite{bahgat2022liwc}.
In our work, we explore static models such as FastText and also state-of-the-art contextual large language models such as BERT and BERTTweet, comparing their performance across evaluation tasks. We closely examine the model's failure cases and hypothesize the potential causes for these failures. 

Many attempts have been made to understand social media language by detecting offense and hate in tweets and conversational data. \cite{2020-semeval} poses a challenge of multilingual hate speech detection in social media.
\cite{autohate} built an automated framework for hate speech detection and separation of hate speech from offensive language. \cite{hatebert} retrains BERT model for abusive language detection in social media. Hate speech and offensive language often contain slang words. We evaluate our models on these tasks to verify if the language models capture slang words and their context. 
\section{Conclusions}

In this paper, we devised a slang understanding framework that combined language models and knowledge sources to help moderators better combat harmful, offensive, or inappropriate content on social media platforms. We applied this framework to three tasks that intuitively rely on slang language, observing that the best performance was obtained by Transformer language models that have been adapted to social media data, such as BERTTweet. Typically, slang usage repositories like Urban Dictionary improved the performance of embedding models that were not adapted to such data before. We found that retraining the models on social media or slang data did not bring consistent gain across tasks. Our error analysis identified five main challenges for slang understanding at scale: incorrect tokenization for infrequent words, presence of URLs, polysemy between formal and slang words, variance in spelling, and misclassifications in the ground truth data. 

Our experiments point to two key aspects that need to be considered seriously in future work. First, Internet slang is an unprecedented form of language evolution that may be solvable by collecting a comprehensive collection of representative data, including memes, slang dictionaries, and usage data. The counterpoint to this data-driven approach is that slang and meme expressions may already be partially designed to avoid censorship, thus anticipating the data-driven solution. Second, as judging the harmfulness of online content is inherently subjective, we suggest a shift from the dominant practice of enforcing a single agreed perspective to the new and emerging practice of modeling disagreement as a signal rather than noise. This would lead to optimizing AI models to make decisions based on the perspectives of individuals or groups, rather than a single best decision.

\section*{Acknowledgements}

The first two authors have been supported by armasuisse Science and Technology, Switzerland under contract No. 8003532866. 

\bibliography{slang}


\end{document}